\documentclass[conference, letterpaper]{IEEEtran}
\IEEEoverridecommandlockouts

\usepackage{cite}
\usepackage{amsmath,amssymb,amsfonts}
\usepackage{algorithmic}
\usepackage{graphicx}
\usepackage{textcomp}
\usepackage{xcolor}

\usepackage{color,soul,cite}
\usepackage{url}
\usepackage{booktabs}
\usepackage{xspace}
\usepackage{rotating}
\usepackage{graphics} 
\usepackage{cite}
\usepackage{amsmath}
\usepackage{amssymb}
\usepackage{subfig}
\usepackage{multirow}
\usepackage{siunitx}

\usepackage{enumitem}
\usepackage{adjustbox}

\newcommand{\ie}{i.e.\xspace}

\newcommand{\rldwa}[1]{3D RL-DWA #1}
\newcommand{\fref}[1]{Fig.~\ref{#1}}
\newcommand{\sref}[1]{Sec.~\ref{#1}}
\newcommand{\eref}[1]{(\ref{#1})}
\newcommand{\tref}[1]{Table~\ref{#1}}

\makeatletter
\let\NAT@parse\undefined
\makeatother
\usepackage[plainpages=false,hypertexnames=true,pdfnewwindow=true,backref=true,colorlinks=true,citecolor=blue,linkcolor=red,urlcolor=blue,filecolor=blue]{hyperref}

\def\BibTeX{{\rm B\kern-.05em{\sc i\kern-.025em b}\kern-.08em
    T\kern-.1667em\lower.7ex\hbox{E}\kern-.125emX}}
\begin{document}

\title{3D RL-DWA: A Hybrid Reinforcement Learning and Dynamic Window Approach for Goal-Directed Local Navigation in Multi-DoF Robots
\thanks{This work was supported by	the  European Union’s Horizon Europe Research and  Innovation  Programme under  Grant  Agreement \#101070066, project REGO.}
}

\author{\IEEEauthorblockN{1\textsuperscript{st} Chiara Castellani}\\
\IEEEauthorblockA{\textit{Humanoids and Human Centered} \\
\textit{Mechatronics} \\
\textit{Istituto Italiano di Tecnologia}\\
16163, Genoa, Italy \\
chiara.castellani@iit.it \vspace{-0.8em}}
\and
\IEEEauthorblockN{2\textsuperscript{nd} Enrico Turco}\\
\IEEEauthorblockA{\textit{Humanoids and Human Centered} \\
\textit{Mechatronics} \\
\textit{Istituto Italiano di Tecnologia}\\
16163, Genoa, Italy \\
enrico.turco@iit.it \vspace{-0.8em}}
\and
\IEEEauthorblockN{3\textsuperscript{rd} Domenico Prattichizzo}\\
\IEEEauthorblockA{\textit{Dept. of Information Engineering and} \\
\textit{Mathematics} \\
\textit{University of Siena}\\
53100, Siena, Italy \\
domenico.prattichizzo@unisi.it \vspace{-0.8em}}
}

\maketitle

\begin{abstract}
	
	In this paper, we present a novel hybrid approach that combines Reinforcement Learning (RL) with Dynamic Window Approach (DWA) for adaptive 3D local navigation of high-degree-of-freedom robotic systems.
	Our method leverages sparse point cloud data to dynamically adjust both the motion and the shape of a deformable microrobot, enabling the system to navigate toward a goal in complex, constrained environments while maximizing the occupied volume.
	We evaluate our framework in a simulated vascular network. 
	Experimental results, based on 1080 trials, indicate that integrating RL with a DWA-based local planner significantly enhances both deformation and navigation capabilities compared to pure RL and model-based methods. In particular, the proposed autonomous controller consistently achieves high deformation and near-perfect path completion during training and maintains robust performance in unseen scenarios. 
	These findings highlight the potential of hybrid planning strategies for efficient and adaptive 3D navigation under sparse sensory conditions.
	
\end{abstract}


\section{Introduction}\label{sec:introduction}
Autonomous navigation is typically achieved using a combination of global and local planning approaches\cite{lixing2023path}. Global path planning computes a trajectory from the start to the goal pose while considering obstacles and robot kinematics. However, these strategies assume a fully known and static environment, making them impractical for real-world applications where conditions are often partially known and dynamically changing\cite{laugier2007autonomous}.
In contrast, local path planning plays a crucial role in enabling robots to adapt in real-time, responding to environmental changes through onboard sensor data (e.g., LiDAR or cameras). This allows for reactive obstacle avoidance and continuous trajectory refinement. 

Navigation in 3D environments is highly complex, especially for robots with high degrees of freedom (DoF)~\cite{Jia2019Fast, kang2024artt}. These systems must efficiently coordinate multiple motion variables while ensuring collision-free navigation.  Moreover, holonomic and external constraints further complicate real-time motion planning, requiring adaptive control strategies.

\begin{figure}
    \centering
	\includegraphics[width=\columnwidth]{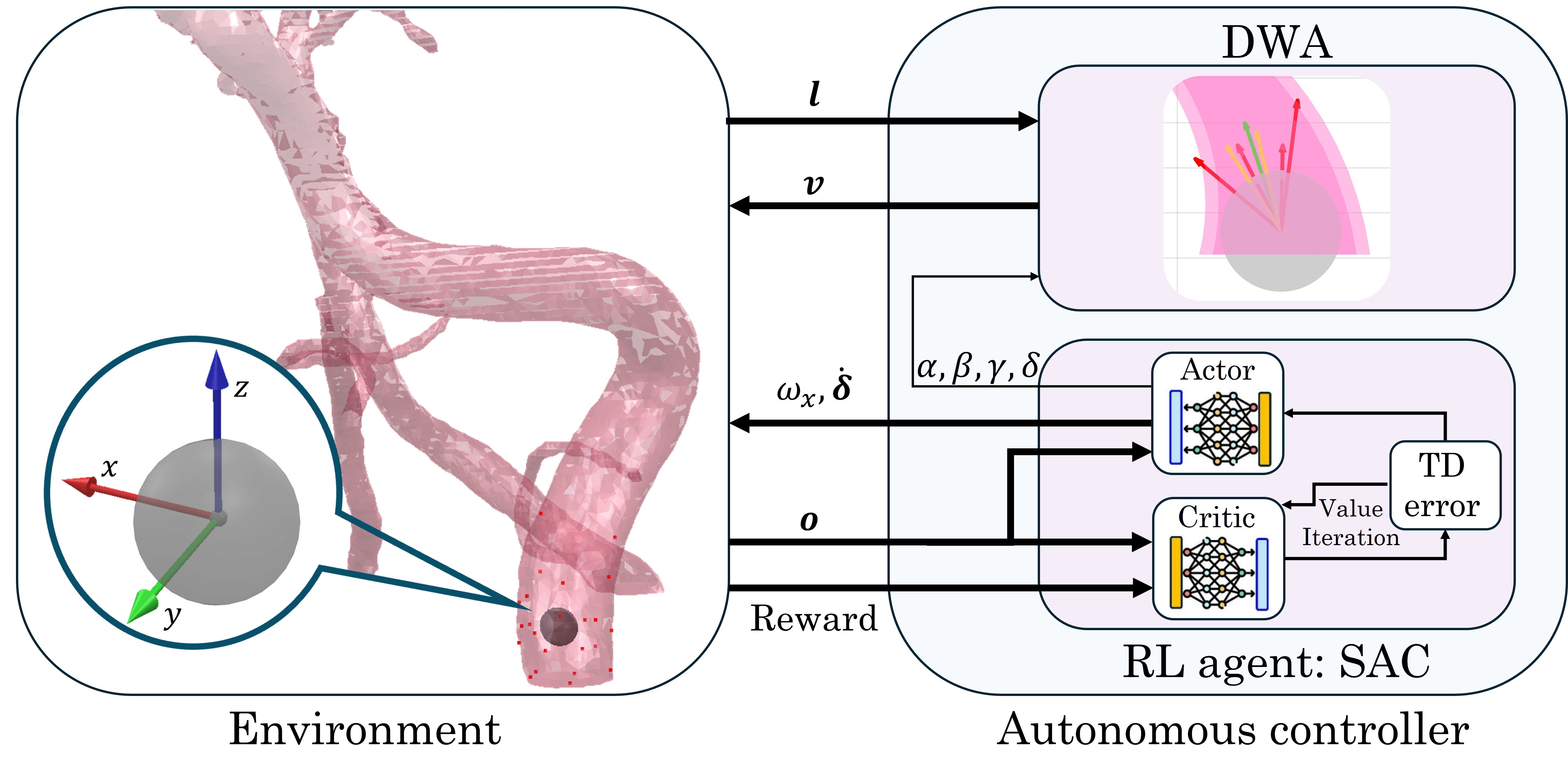}
    \caption{Proposed framework for adaptive 3D local navigation.
   	The autonomous controller combines RL and DWA within a closed-loop system. At each time step, the controller receives an observation vector $\boldsymbol{o}=[\boldsymbol{s},\boldsymbol{l}]$ from the environment, where $\boldsymbol{s}$ denotes the current task state and $\boldsymbol{l}$ contains the laser-scan readings.
   	The RL agent dynamically adjusts the DWA cost-function weights ($\alpha, \beta, \gamma, \zeta$) and outputs angular $\omega_x$ and deformation velocities $\boldsymbol{\dot{\delta}}$. The DWA module then selects the optimal linear velocity $\boldsymbol{v}$ from the admissible velocity set.}
    
    \label{fig:block_diagram}
\end{figure}
    
One of the most widely used local navigation techniques is the Dynamic Window Approach (DWA)~\cite{Fox1997DWA}. This method optimizes the planned path by evaluating a cost function that balances factors such as distance to obstacles and alignment with the goal. While effective, conventional DWA has limitations, particularly its reliance on fixed parameter tuning, which makes adaptation to dynamic environments difficult. 
Traditional DWA requires extensive manual tuning of cost function weights to adapt to different scenarios, such as navigating narrow gaps or open spaces. This limits adaptability in dynamic environments and increases computational costs in high-dimensional motion
spaces, complicating real-time motion planning.

Recent advancements in Reinforcement Learning (RL) have shown promise in adaptive navigation, enabling robots to dynamically adjust their control policies based on environmental conditions~\cite{zhu2021deep}. Hybrid approaches combining RL with classical planners, such as DWA~\cite{Kim2022Improvement, liu2024td3basedcollisionfree, Dobrevski2024Dynamic}, have demonstrated improved adaptability and robustness. 
However, existing work has primarily focused on 2D motion planning in structured environments, leaving a gap in solutions for 3D motion control in high-DoF robotic systems.

In this work, we extend these emerging approaches and propose a novel framework, namely the \rldwa (shown in \fref{fig:block_diagram}), for the navigation of a 9-DoF deformable robot. 
Our system combines RL-driven decision-making with a DWA-based local control, allowing the robot to dynamically adjust its linear, angular, and deformation velocities. We specifically address goal-directed local navigation, where the global route is defined as a 
limited sequence of waypoints. 
Our contribution is an adaptive local controller that guides the robot toward each waypoint while simultaneously optimizing its deformation, leveraging sparse sensory data.

The robot is required to navigate in a complex and constrained environment, consisting of a simulated vascular network, toward a selected target while maximizing the occupied volume and avoiding collisions with the vessels.
%
Robot actuation is described through a simplified kinematic model that captures both motion and deformation, and is inspired by recent advances in ferrofluidic robotic systems~\cite{Xinjian2023Combined, ji2022Deformable, Yang2022Review}. However, in this work, we focus on high-level control and do not explicitly model magnetic forces and material dynamics.
The effectiveness of our approach is validated through extensive experimental trials, 
where we compare performance against a pure RL and a model-based navigation method.


\section{Related Works}\label{sec:related_works}
Research on autonomous navigation has traditionally focused on 2D motion for mobile robots, extending to 3D applications in domains such as unmanned aerial vehicles 
(UAVs) and underwater robots~\cite{nahavandi2025comprehensive}. 
However, as robotic systems operate in increasingly complex and dynamic environments, there is a need for advanced navigation strategies with adaptive decision-making and real-time reconfigurability. In such scenarios, local path planning plays a crucial role in ensuring efficient and collision-free navigation.
%

Among local path planning techniques, artificial potential field, optimization-based methods, and learning-based strategies have been widely employed~\cite{lixing2023path, Romero2024Actor}. 
In particular, DWA has proven highly effective in dynamic environments, as it optimizes velocity profiles through a cost function that balances target heading, obstacle avoidance, and motion continuity.
Extensions of traditional DWA to 3D applications were explored in UAVs~\cite{Jorge2024DWA3D} and underwater robots~\cite{Tusseyeva20133D}.
However, DWA relies on fixed parameter tuning, which limits adaptability in dynamic and unstructured environments and often requires extensive experimentation~\cite{Jorge2024DWA3D}.
To address this limitation, several works have explored hybrid approaches that integrate adaptive parameter tuning into DWA-based navigation. Neuro-fuzzy and fuzzy-logic 
approaches have been used to adjust the DWA parameters online according to environmental conditions~\cite{teso2019predictive,Abubakr2022intelligent,Xiang2022parameter}.
In parallel, RL techniques have been integrated  with DWA to improve adaptability. 
In~\cite{Chuanbo2023deep}, DWA was used to refine reward signals in a Proximal Policy Optimization (PPO) RL framework, while~\cite{Kim2022Improvement} introduced an RL-based controller to fine-tune the DWA velocity outputs. 
Further developments have included the TD3-DWA approach~\cite{liu2024td3basedcollisionfree} for collision-free motion planning, and DWA-RL~\cite{Patel2020DWARLDF}, which integrates deep RL with DWA to improve navigation among mobile obstacles.
Additionally, Dobrevski and Skočaj~\cite{Dobrevski2024Dynamic} proposed an RL framework that leverages high-dimensional laser scan observations to dynamically adjust DWA’s parameters, while in~\cite{Chang2021reinforcement}, Q-learning has been employed for DWA parameter optimization.
Despite these advancements, most RL-DWA methods remain focused on 2D navigation~\cite{Dobrevski2024Dynamic,Patel2020DWARLDF,Kim2022Improvement,liu2024td3basedcollisionfree,Chang2021reinforcement} and do not address high-DoF systems operating in constrained 3D spaces.

Other hybrid methods integrate RL with Model Predictive Control (MPC). 
Romero et al.~\cite{Romero2024Actor} proposed an Actor-Critic MPC architecture for quadrotors that embeds a differentiable MPC layer within the actor network.
More recently, Ramezani et al.~\cite{ramezani2023uav} introduced an LSTM-based MPC integrated into a DDPG framework for UAV path planning, while Martinsen et al.~\cite{martinsen2022reinforcement} proposed an RL-based MPC controller for unmanned surface vessels. 
While promising, these methods involve computationally demanding online optimization and rely on accurate dynamical models.
	
In this work, we instead adopt DWA, as it provides a lightweight sampling-based strategy that is particularly suited to real-time navigation in constrained environments. This problem is especially relevant in emerging domains such as the autonomous control of microrobots for biomedical applications. Recent advances in microrobotics have introduced high-DoF platforms, such as adaptive microparticle swarms~\cite{yu2022adaptive} and deformable ferrofluidic microrobots~\cite{ji2022Deformable,Xinjian2023Combined}, targeting applications such as minimally invasive interventions and precision drug delivery.
In this context agents must navigate highly constrained and unstructured environments in conditions of sparse exteroceptive data. 
Although efforts have been made to develop adaptive control strategies for autonomous navigation~\cite{Yang2022Hierarchical,Liu2025Autonomous}, 
their applicability to high-dimensional systems in constrained 3D environments remains limited.
%
%
%
Our method addresses this gap by controlling a high-DoF microrobot in constrained 3D environments, leveraging sparse sensory data to optimize both motion and shape.

\section{Materials and Methods}\label{sec:methodology}
\subsection{System Overview} \label{sec:system_overview}

The system models the motion of a deformable robot navigating autonomously through a constrained and tortuous environment consisting of a reproduction of blood vessels~\cite{sketchfab_blood_vessel}.

The robot is modeled as a deformable ellipsoidal system, initially arranged in a spherical shape. 
It features 9 DoF: 6 DoF define its pose (including its position $\boldsymbol{p}=[x,y,z]$ and orientation $\boldsymbol{q}=[q_x,q_y,q_z,q_w]$), while 3 additional DoF allow modifying the shape of the robot along its three axes of symmetry ($\boldsymbol{\delta}=[\delta_x, \delta_y, \delta_z]$).
\fref{fig:swarm_def} shows how the deformations occur.
The robot is equipped with $N$ laser sensors, positioned at its center and evenly distributed along a spherical configuration. Varying $N$ allows us to reproduce contexts with different levels of exteroceptive information.
The sensors remain fixed in the robot’s reference frame, maintaining a constant relative position and orientation.

\begin{figure}
	\vspace{4pt}
    \centering
{\includegraphics[width=0.7\columnwidth]{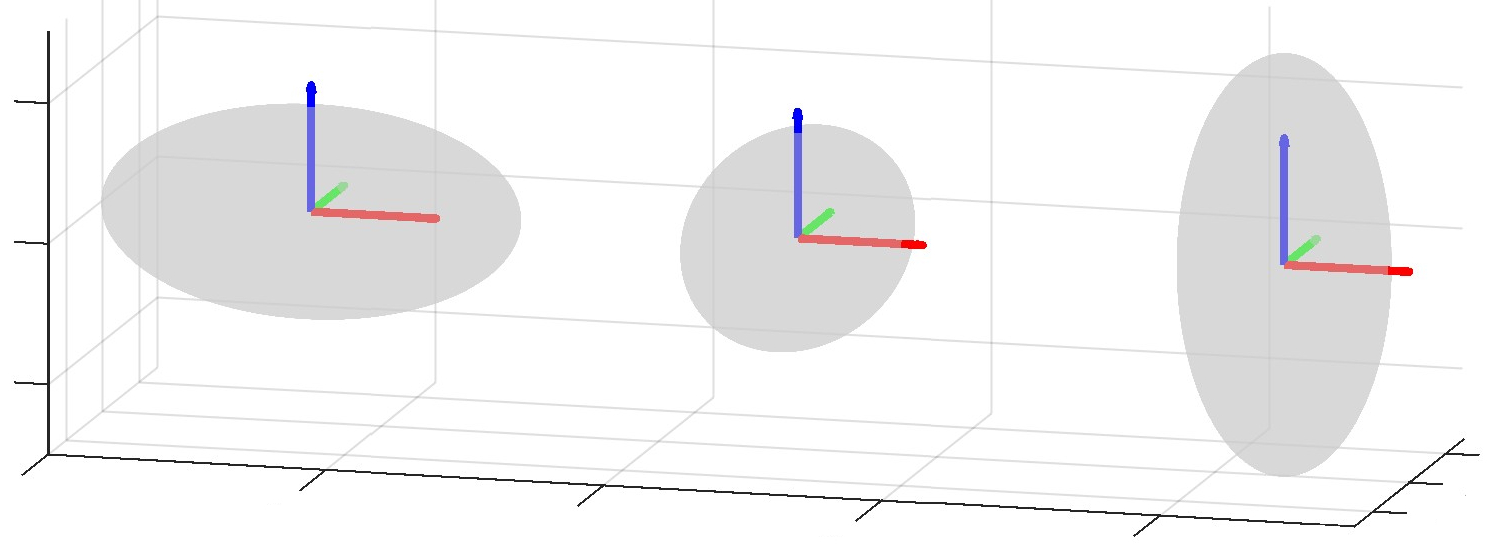}\label{fig:swarm_def}}
    \caption{Deformations of the robot along the three axes of its reference system. The axes are represented in red ($x$), green ($y$), and blue ($z$).}
    \label{fig:swarm_def}
\end{figure}

\subsection{Robot Kinematics}

In our approach, the robot is velocity-controlled. 
Let $\boldsymbol{v}_t \in \mathbb{R}^3$ and $\dot{\boldsymbol{\delta}}_t \in \mathbb{R}^3$ denote the commanded translational and deformation velocities at time step $t$, respectively.
Assuming constant velocities over each control interval $\Delta t$, we express the translational and deformation equations as:
\begin{equation}
	\label{eq:kinematicmotion}
	\begin{aligned}
		\boldsymbol{p}_{t+1} &= \boldsymbol{p}_{t} + \boldsymbol{v}_{t} \Delta t, \\		\boldsymbol{\delta}_{t+1} &= \boldsymbol{\delta}_{t} + \dot{\boldsymbol{\delta}}_{t} \Delta t. 
	\end{aligned}
\end{equation}
Although the system features 9 DoF, as described in~\sref{sec:system_overview}, 
we directly control only a subset of them. Specifically, we imposed that the robot's $x$-axis aligns with the direction of motion. 
Consequently, rotations about the $y$ and $z$ axes are constrained by this alignment, leaving the rotation about the $x$-axis as the sole controllable rotational DoF. 
This design choice is inspired by magnetically actuated robots, where external torques inherently align the system with the 
motion direction due to the applied magnetic field constraints~\cite{Liu2025Autonomous}.

The robot's angular velocity $\boldsymbol{\omega}_t$ consists of two components: (\textit{i}) a passive component $\hat{\boldsymbol{\omega}}_t$ induced by the alignment constraint, and (\textit{ii}) an actively controlled component $\boldsymbol{\omega}_{ctrl}=[\omega_{x},0,0]$.
Let $R_t \in SO(3)$ be the rotation matrix associated with the orientation quaternion $\boldsymbol{q}_t$, and let $\hat{R}_{t+1} \in SO(3)$ be the constrained orientation whose body-frame $x$-axis is aligned with the direction of motion. 
The relative rotation from the current orientation $R_t$ to the constrained orientation $\hat{R}_{t+1}$ is defined as:
\begin{equation}
	\label{eq:delta}
	\Delta R_t = \hat{R}_{t+1} R_t^\top.
\end{equation}
%

%
The passive angular velocity component is approximated 
from the skew-symmetric part of $\Delta R_t$ as:
\begin{equation}
	\label{eq:passive_omega}
	\hat{\boldsymbol{\omega}}_t =
	\frac{1}{2\Delta t}
	\begin{bmatrix}
		\Delta R_{t,3,2} - \Delta R_{t,2,3} \\
		\Delta R_{t,1,3} - \Delta R_{t,3,1} \\
		\Delta R_{t,2,1} - \Delta R_{t,1,2}
	\end{bmatrix},
\end{equation}
where $\Delta R_{t,i,j}$ denotes the $(i,j)$ element of $\Delta R_t$. 
Eventually, we obtain the quaternion update by integrating the angular velocity $\boldsymbol{\omega}_t$ over the time step $\Delta t$. The incremental quaternion rotation follows from the matrix exponential:
\begin{equation} \boldsymbol{q}_{t+1} = \boldsymbol{q}_t \cdot e^{\frac{1}{2} S(\boldsymbol{\hat{\omega}}_t+\boldsymbol{\omega}_{ctrl}) \Delta t}, \end{equation}
where $S(\cdot)$ is the skew-symmetric matrix.

\subsection{Control Architecture}

 \begin{figure}
 		\vspace{4pt}
	    \centering
	      \includegraphics[height=4.5cm]{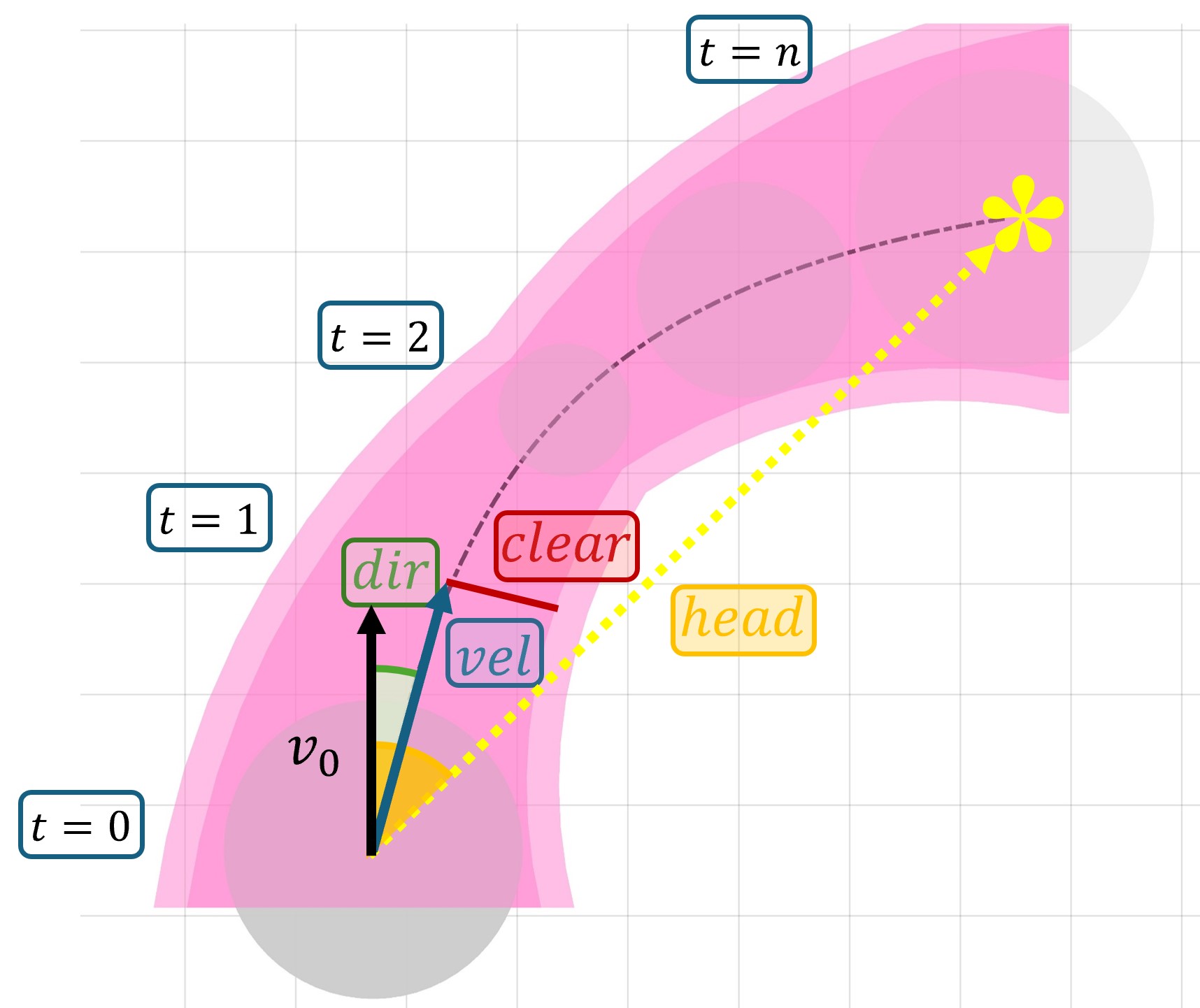}
	    \caption{Visualization of the components of the DWA cost function: \textit{vel} (blue), \textit{dir} (green), \textit{clear} (red), and \textit{head} (orange), evaluated based on the current robot velocity $v_0$ during navigation toward a target (yellow). The robot is navigating a deformed curved path at time steps \(t=0, 1, 2, \ldots, n\). }
	    \label{fig:path_dwa}
\end{figure}
Our control architecture for local path planning combines a customized 3D DWA method with an RL module to achieve adaptive path planning and deformation control while navigating toward a target.
\fref{fig:block_diagram} provides an overview of the entire framework. 
The RL agent processes observation $\boldsymbol{o}$ from the environment (i.e., task state $\boldsymbol{s}$ and laser scan data $\boldsymbol{l}$) and outputs the DWA parameters ($\alpha, \beta, \gamma, \zeta$) and the robot’s angular and deformation velocities ($\omega_x$, $\boldsymbol{\dot{\delta}}$ ). The DWA then determines the optimal linear velocity accordingly. 

\subsubsection{3D Dynamic Window Approach}

We designed a customized 3D DWA method tailored to our specific task and the robot’s kinematics. Specifically, the planner extends the standard DWA formulation to 3D translational motion and selects admissible velocity commands that balance obstacle avoidance, motion efficiency, and progress toward the selected target.
The method first computes a set of admissible velocities, which form the Dynamic Window $V_d$. These velocities are determined by considering the robot’s current velocity $\boldsymbol{v_t}$ and its maximum acceleration along each axis $\boldsymbol{\dot{v}}^{\text{max}}$:
\begin{equation}
	\label{eq:velocity_window}
	\begin{aligned}
			 V_d = \{ \boldsymbol{v} \,\big|\, 
			&\boldsymbol{v} \in \left[\boldsymbol{v_t} - \boldsymbol{\dot{v}^{\text{max}}} \cdot \Delta t, \, \boldsymbol{v_t} + \boldsymbol{\dot{v}^{\text{max}}} \cdot \Delta t\right] \}.
		\end{aligned}
\end{equation}

From this set, the optimal velocity is selected as the one that maximizes the cost function $G$:
\begin{equation}
\label{eq:dwa_costfunction}
    G(\boldsymbol{v}) = \alpha \cdot vel(\boldsymbol{v}) + \beta \cdot dir(\boldsymbol{v}) + \gamma \cdot clear(\boldsymbol{v}) + \zeta \cdot head(\boldsymbol{v}).
\end{equation}
The DWA factors are visually rendered in \fref{fig:path_dwa}. In our formulation, all terms are normalized to ensure comparable scaling.
Specifically, $vel(\boldsymbol{v})$ assigns a cost based on the translational speed, penalizing lower velocities to encourage faster motion, 
while $dir(\boldsymbol{v})$ measures the angular deviation between the candidate and the robot's current velocity, penalizing rapid changes in direction such as backward motions.
The function $clear(\boldsymbol{v})$, representing clearance, evaluates the minimum distance to the nearest obstacle along the predicted trajectory, penalizing velocities that bring the robot too close to obstacles. The trajectory is computed based on the robot's current velocity, and it is discretized with step $\Delta t$  within a fixed time horizon $T$. 
Finally, $head(\boldsymbol{v})$ measures the deviation between the robot's current motion direction and the target direction, encouraging alignment of the heading toward the goal.
%
The weights $\alpha$, $\beta$, $\gamma$, and $\zeta$ allow the controller to prioritize speed, smooth direction changes, obstacle avoidance, and goal alignment as needed.
Thus, robot behavior can be adapted to different scenarios by varying these parameters, as the optimal local trajectory will vary accordingly. 
After selecting the best velocity from $V_d$, the robot’s trajectory is properly updated. 
%

\subsubsection{Reinforcement Learning Module}

The RL module formulates the navigation problem as a Markov Decision Process (MDP), where the agent learns to optimize the parameters of the DWA cost function \eref{eq:dwa_costfunction} and compute optimal rotational and deformation velocity commands. The objective is to achieve adaptive 3D local navigation while maximizing the volume occupied by the robot. The task is formulated so that the learned policy must co-optimize pose and shape, thereby enforcing simultaneous control of all DoF. 
The key components of the MDP are defined as follows:

\begin{itemize}   
    \item \textbf{Observation ($\boldsymbol{o}_t$)}: represents the robot's current understanding of the task. It consists of the current task state $\boldsymbol{s}_t$ (\ie, the robot's deformation $\boldsymbol{\delta}_t$, the normalized current velocity $\frac{\boldsymbol{v}_t}{\Vert \boldsymbol{v}_t\Vert }$, and the normalized displacement vector $\boldsymbol{d}_{t}$ pointing from the robot’s current position toward the goal), and the $N$ distances to obstacles represented as $\boldsymbol{l}_t$, extracted from the laser scan data.
    \item \textbf{Action ($\boldsymbol{a}_t$)}: at each step, the agent outputs an action $\boldsymbol{a}_t$ 
    that includes the DWA weighting factors ($\alpha$, $\beta$, $\gamma$, $\zeta$, ranging within $[0, 1])$, the rotational velocity around the robot's $x$-axis 
    (${\omega}_x$), and the deformation velocities values along the three principal axes $\dot{\boldsymbol{\delta}}_t = [\dot{\delta}_x, \dot{\delta}_y, \dot{\delta}_z]$.

     \item \textbf{Reward ($R_t$)}: we define a dense reward function to promote effective local navigation. 
     The reward at each time step is computed as:
    \begin{equation}
    R_t = R_{\text{def}} + R_{\text{stuck}} + R_{\text{outside}} + R_{\text{goal}},
    \label{eq:reward}
    \end{equation}
    where $R_\text{def}$ encourages the robot to maximize its deformation 
    while ensuring no collisions occur. The deformation reward is defined as follows: 
     \begin{equation}
         \begin{cases} 
             \sqrt[3]{\delta_x \cdot \delta_y \cdot \delta_z} & \text{if no collision occurs}, \\
             0 & \text{otherwise}.
         \end{cases}
     \end{equation}
    $R_\text{stuck}$ penalizes the robot when stuck (no significant progress in the last 20 steps) or moving backwards, while $R_\text{outside}$ assigns a penalty if the robot exits the designated navigation space. Note that the blood vessel walls are virtual, and the robot can exit the path at any point. Lastly, $R_\text{goal}$ assigns a positive reward when the target is successfully reached. The values of $R_{stuck}$, $R_{outside}$, and $R_{goal}$ are constant and set the magnitude of the corresponding penalty or bonus.


\end{itemize}
For our reinforcement learning module, we adopted the Soft Actor-Critic (SAC) algorithm due to its sample efficiency, robust convergence, and stability.
We compared the performance of SAC with a 
PPO framework~\cite{ppo} and found that SAC consistently achieved higher convergence rates and greater learning stability.

\subsection{Evaluation Metrics}
\label{sec:metrics}
To evaluate our approach, we defined two performance metrics: the path completion percentage $p_\%$ and the percentage of the occupied volume $V_{\%}$. The former measures the percentage of the total length successfully traversed by the robot at the end of each episode. The latter, instead, quantifies how effectively the robot adapts its shape to the vascular structure. At each step, we compute the ratio between the robot’s occupied volume and the volume of the currently observed vessel segment; the final metric is obtained by averaging these values over the entire episode.
When the robot collides with the vessel walls, this metric is set to zero.

\section{Experimental Evaluation}\label{sec:experiments}
\begin{figure}
	\vspace{4pt}
	\centering
	\includegraphics[width=0.78\columnwidth]{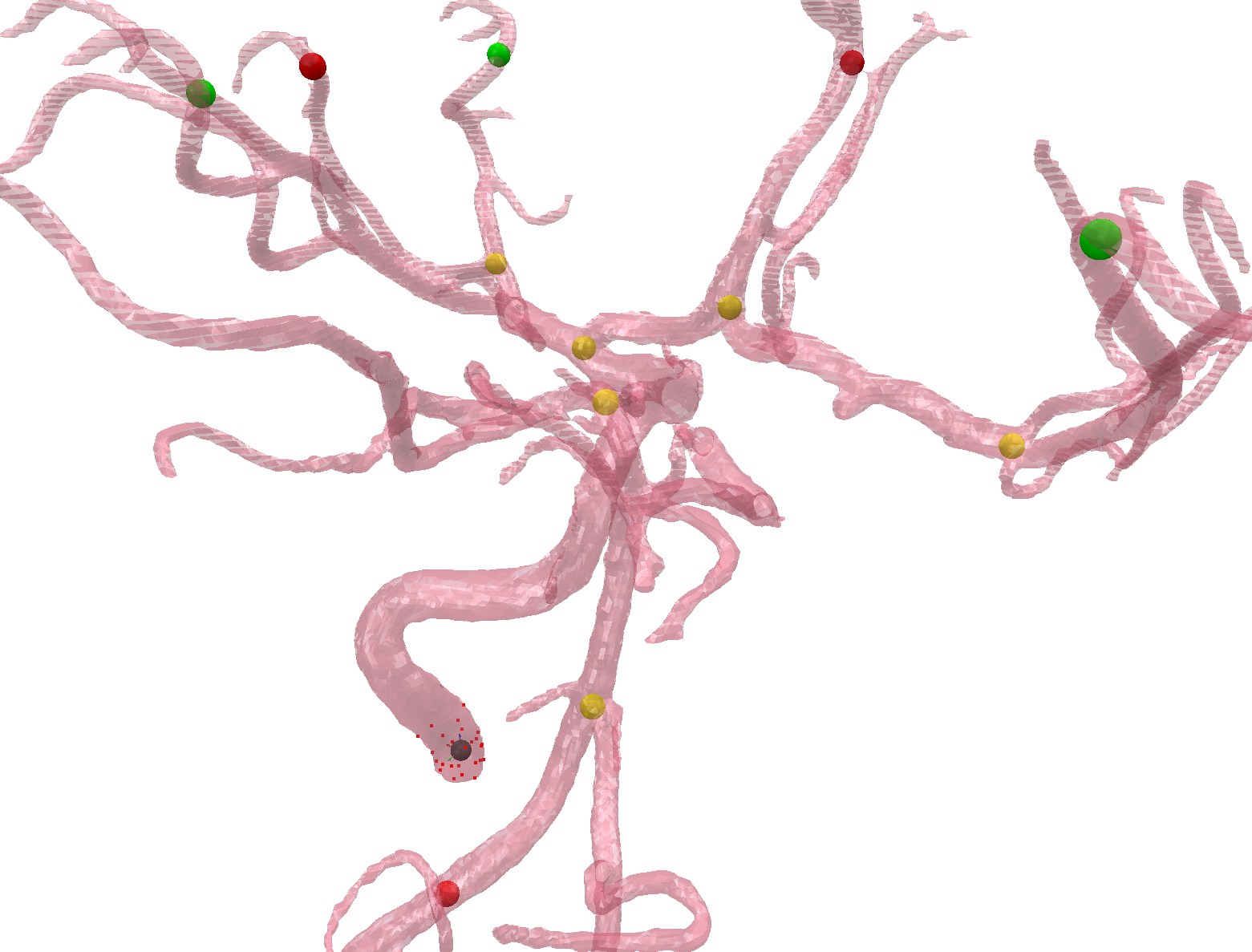}
	\caption{Training and test scenario. Yellow spheres placed at junctions of the network represent sparse navigation waypoints. Final target locations used for training are shown as green spheres, while test locations are shown as red spheres. }
	\label{fig:mazes}
\end{figure}

We evaluated our \rldwa approach in a simulated environment reproducing an intricate vascular network (see \fref{fig:mazes}). The robot was initialized at a fixed starting position and had to navigate through sequences of sparse navigation targets placed at critical junctions of the network, continuously adapting its local motion and shape to maximize the reward while avoiding exiting from the vessel lumen. The overall global trajectory emerged as the concatenation of $w$ successive local goals. This experimental design allowed us to assess not only the robot’s ability to complete individual trajectories, but also its capacity to generalize across unseen waypoint sequences within the same complex environment.

To investigate the impact of sensing resolution on learning and generalization, we trained our model using three input sizes for the scan readings ($N=15, 25, 50$).
We tested the system on both training and unseen target sequences, and compared our controller against two baseline methods.
The first is an 
RL SAC approach, which 
predicts linear, angular, and deformation velocities, and is used to 
highlight the benefits of integrating the DWA module to enhance motion learning and improve local navigation efficiency.
The second is a model-based controller~\cite{turco2024reducing}, 
which computes the Delaunay triangulation of the point cloud, and constructs a dual Voronoi graph. The shortest Euclidean path on this graph, spanning from the start to the end of the detected vessel section, defines the optimal trajectory. 
Optimal deformation values are extracted from the vessel cross-sectional geometry along the path.

%

In addition, two further analyses were performed. 
First, we measured inference times and analyzed the computational complexity of the proposed method to evaluate the real-time performance of our controller.
Second, to assess robustness under sensor uncertainties, we injected Gaussian noise of increasing magnitude 
into the laser scan data and systematically evaluated performance 
across all sensing resolutions.


\subsection{Training}

At the start of each training episode, a sequence of intermediate targets was randomly selected to guide the robot toward one of the three 
final locations, highlighted in green in Fig.~\ref{fig:mazes}. All 
trajectories from the start to the final target had comparable length and the same number of waypoints ($w=3$). 

The episode reward is computed as the sum of the step-wise rewards \eqref{eq:reward}, which implicitly encourage forward motion, collision avoidance, effective deformation, and convergence toward the target. An episode was considered successful if the robot reached the selected target, 
while it was terminated if the robot exited the vessel lumen or failed to progress forward for more than 50 consecutive steps.

To optimize training, we performed a grid search over a set of relevant hyperparameters. 
For each 
configuration, we ran 5 training sessions with different random seeds and evaluated both the loss trends and the reward curves. The optimal hyperparameter values, highlighted in bold in Table~\ref{tab:hyperparameters}, were chosen based on their convergence speed and stability.


\begin{table}[h]
    \centering
    \caption{Grid Search Hyperparameter Space.}
    \begin{tabular}{l c}  
        \toprule
        \textbf{Parameter} & \textbf{Values} \\
        \midrule
        Learning Rate & $\boldsymbol{1e{-}3}, 1e{-}4$ \\
        Discount Factor ($\gamma$) & 0.9, $\boldsymbol{0.92}$, 0.99 \\
        Target Smoothing Coefficient ($\tau$) & $\boldsymbol{0.002}, 0.005$ \\
        Entropy Coefficient & \texttt{auto}, $0.1$, $\boldsymbol{0.2}$ \\
        Batch Size & $\boldsymbol{256}, 512$ \\
        \bottomrule
    \end{tabular}
    \label{tab:hyperparameters}
\end{table}

\begin{figure}
    \centering
	\includegraphics[width=0.98\columnwidth]{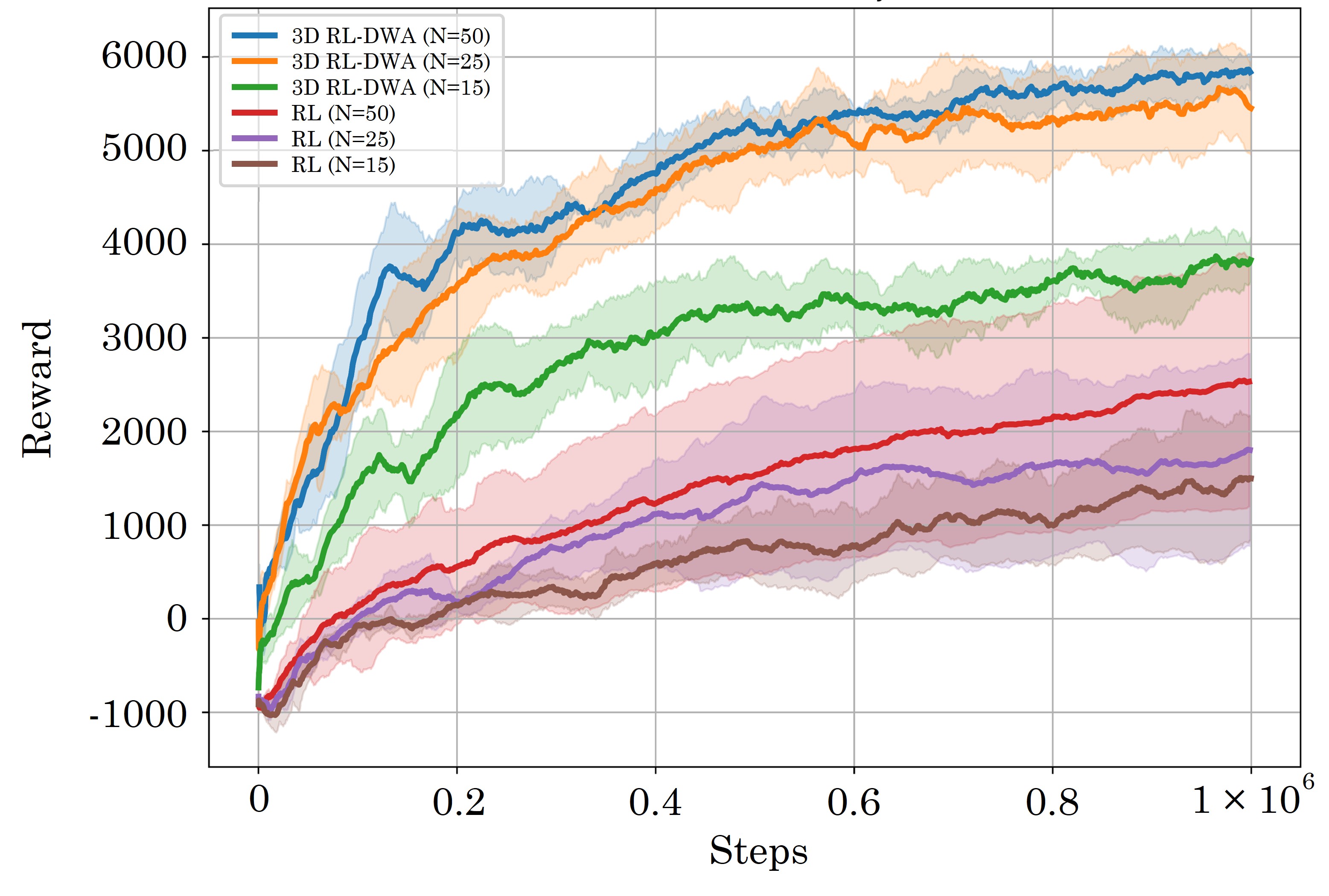}
    \caption{Learning curves of each control strategy, showing the mean episodic rewards. 
	The solid lines indicate the mean results across $5$ trials and the shaded areas are the standard deviations.}
    \label{fig:curves}
\end{figure}

We trained the \rldwa using the three input resolutions and additionally trained the RL SAC baseline. 
Training continued until convergence, which was achieved in all cases within $\num{1e6}$ steps. The resulting learning curves are presented in Fig.~\ref{fig:curves}. To 
complement the qualitative analysis of the learning dynamics, we 
computed the area under the curve (AUC) as a quantitative metric. The AUC provides a compact measure of both the overall learning progress and the speed of reward accumulation throughout training.

The training procedure, implemented in TensorFlow with PyRep for simulation interfacing, lasted approximately 3 hours on a desktop PC equipped with an NVIDIA GeForce RTX 3060 and an Intel Core i7-12700KF. 

\subsection{Test}
\begin{figure}
 	\vspace{2pt}
    \centering
    \subfloat[]{\includegraphics[width=0.85\columnwidth]{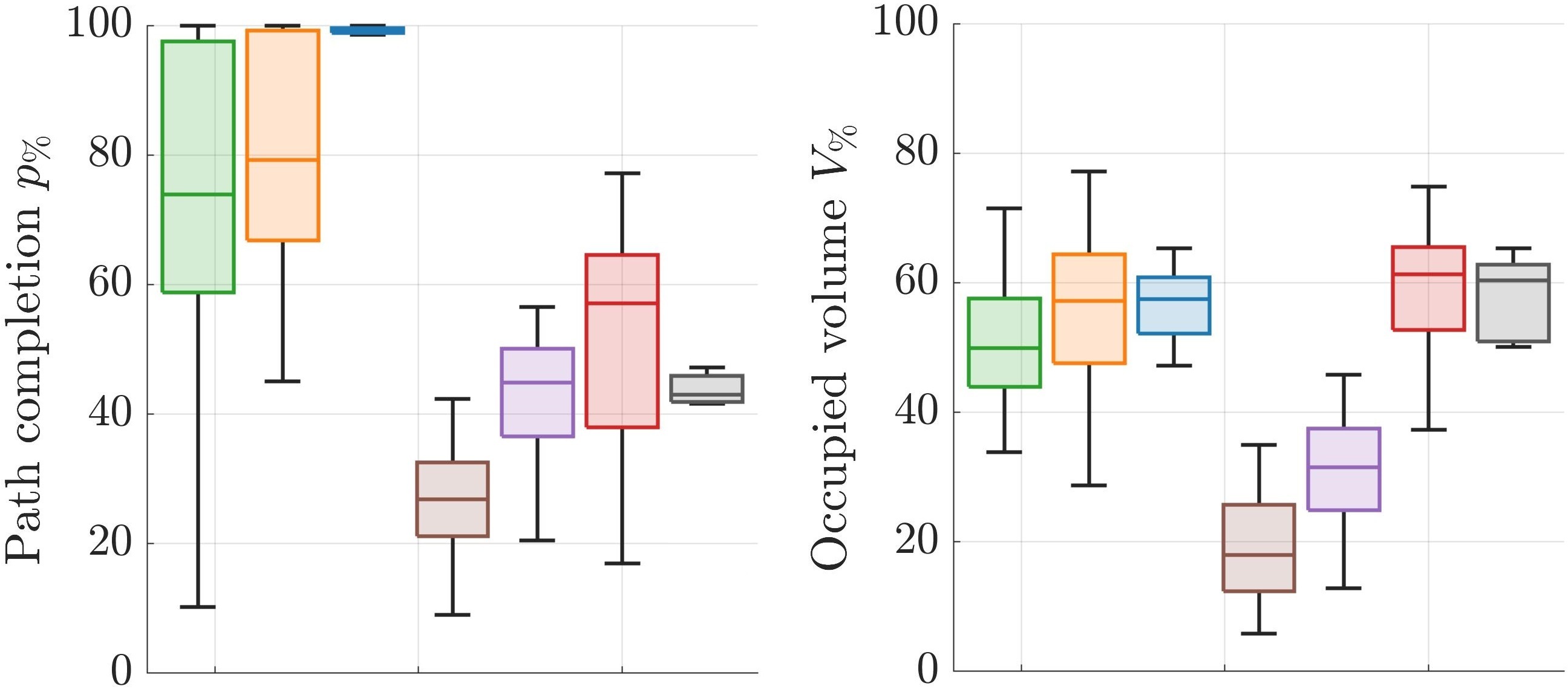}\label{fig:training_res}} \\
    \subfloat[]{\includegraphics[width=0.85\columnwidth]{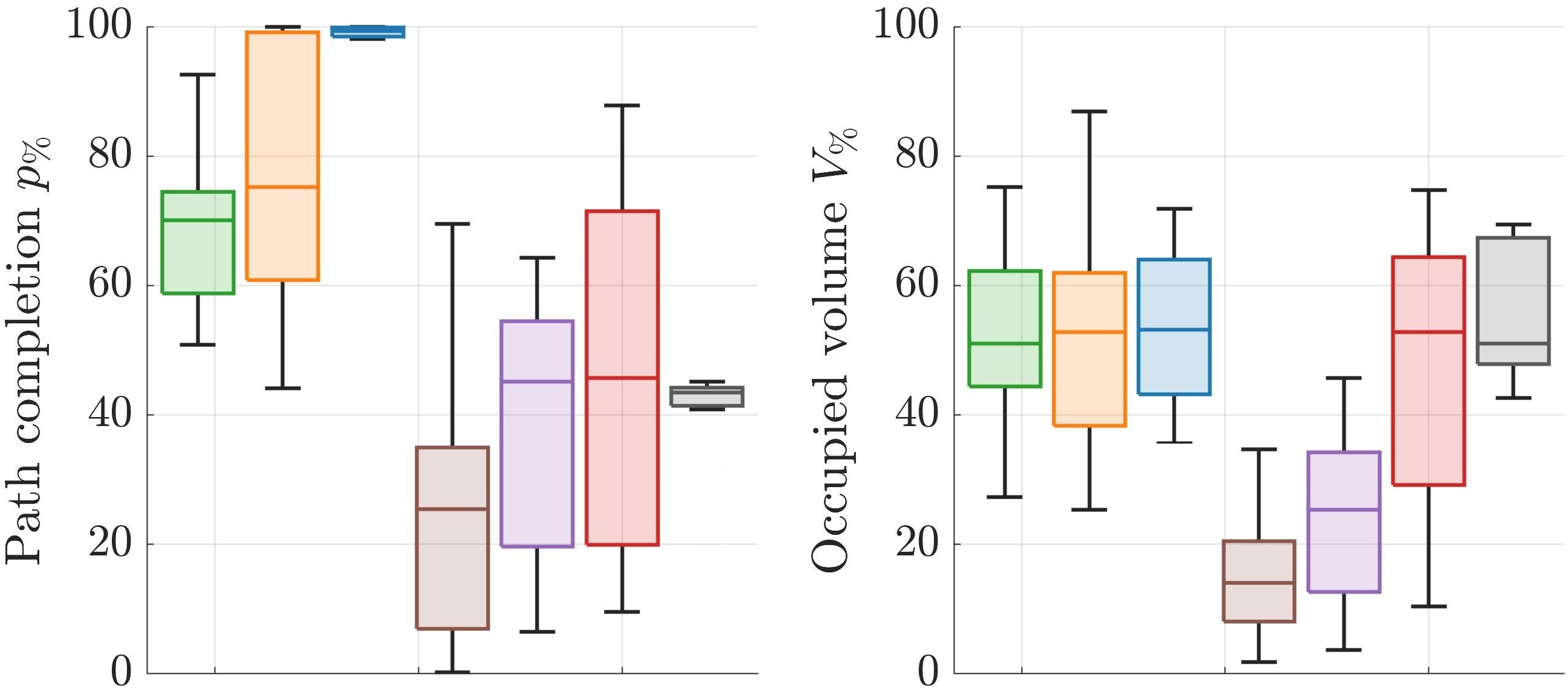}\label{fig:test_res}} 
        	 	   \vspace{0.15cm}
\includegraphics[width=0.9\columnwidth]{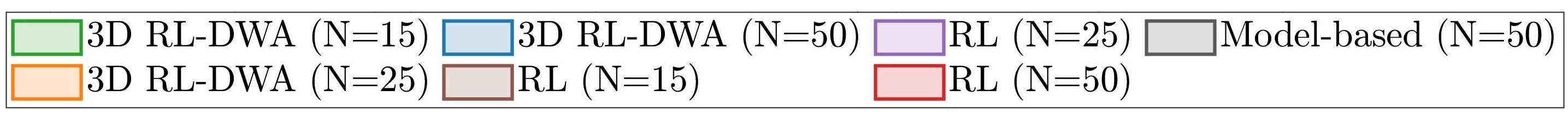}
    \caption{Comparison of quantitative metrics. Median values and interquartile ranges of the control strategies are shown for (a) training scenarios and (b) test scenarios.}
    \label{fig:res}
\end{figure}

The testing phase was conducted both on training and unseen target sequences, guiding the robot toward one of the six final locations highlighted in green (training) and red (test) in Fig.~\ref{fig:mazes}. All sequences share the same number of waypoints $w$ and have comparable lengths. For each path, we conducted 20 trials and recorded the performance metrics described in~\sref{sec:metrics}.
To further evaluate the effectiveness of our approach, particularly in conditions with sparse exteroceptive data, we compared it against the RL baseline, and the model-based method using $N=15$, $25$ and $50$ sensing points. 
Overall, we ran a total of 1080 experimental trials ($20$ repetitions $\times$ $6$ goal sequences $\times$ $3$ control strategies $\times$ $3$ sensor resolutions). However, the model-based approach consistently failed in all trials with $N=15$ and $25$ sensor readings, as the autonomous controller was unable to construct a reliable graph and extract relevant information for local navigation. The results are summarized in~\fref{fig:res}. Note that the results for the model-based approach at lower sensing resolutions are omitted, since it consistently achieved zero-score in all metrics. 

\subsection{Inference time}

To evaluate the computational efficiency, 
we measured the mean inference time per control step and its standard deviation at different sensing resolutions. The analysis separates the contributions of the RL and DWA components. 
Results are reported in \tref{tab:inference_times}.

\subsection{Robustness under uncertainties}

To assess the robustness of the proposed \rldwa approach under sensor uncertainties, we conducted an additional experiment in which Gaussian noise with varying standard deviation $\sigma$ was injected into the sensor readings. Three noise levels were considered: low ($\sigma \approx 5\%$ of the mean laser scan distance in a baseline episode), moderate ($\sigma \approx 10\%$), and high ($\sigma \approx 20\%$). 
For each configuration, the controller was evaluated 20 times across all training and test sequences, 
using input resolutions of $N=15, 25,$ and $50$ sensing points. \tref{tab:exp_robustness} summarizes the results, reporting the mean and standard deviation of the two performance metrics, namely path completion and occupied volume.

\begin{table}[t]
	\centering
	\caption{Mean and standard deviations of inference time.}
	\renewcommand{\arraystretch}{1.15}
	\setlength{\tabcolsep}{8pt}
	\begin{tabular}{l cc}
		\toprule
\multirow{2}{*}{\textbf{\shortstack[l]{Sensor\\readings}}}
		& \multicolumn{2}{c}{\textbf{Inference Time [ms]}} \\
		\cmidrule(lr){2-3}
		& \textbf{RL} & \textbf{DWA} \\
		\midrule
		$N=15$ & $0.43064 \pm 0.00004$ & $1.20672 \pm 0.0004$ \\
		$N=25$ & $0.43037 \pm 0.00012$ & $1.25589 \pm 0.00041$ \\
		$N=50$ & $0.42677 \pm 0.00009$ & $1.43311 \pm 0.0004$ \\
		\bottomrule
	\end{tabular}
	\label{tab:inference_times}
\end{table}


\begin{table*}[t]
	\vspace{4pt}
    \centering
    \caption{Mean path completion percentage and occupied volume per step under different noise levels.}
    \label{tab:exp_robustness}
    \renewcommand{\arraystretch}{1.1}
    \begin{tabular}{l c c c c c c}
        \toprule
        \multirow{2}{*}{\textbf{Method}} & \multicolumn{3}{c}{\textbf{Path completion [\%]}} & \multicolumn{3}{c}{\textbf{Occupied volume [\%]}} \\
        \cmidrule(lr){2-4} \cmidrule(lr){5-7} \cmidrule(lr){6-7}
        & \textbf{Low noise} & \textbf{Medium noise} & \textbf{High noise} & \textbf{Low noise} & \textbf{Medium noise} & \textbf{High noise} \\
        \midrule
       \rldwa (N=15)& $72.1\pm22.0$ & $64.9\pm26.6$ & $51.2\pm27.1$ & $49.7\pm14.2$ & $47.9\pm14.1$ & $46.4\pm14.5$  \\
        \rldwa (N=25)& $78.4\pm19.0$  & $76.2\pm18.7$ & $66.8\pm13.1$ & $57.9\pm13.6$ & $56.4\pm13.9$ & $55.1\pm13.6$  \\
        \rldwa (N=50)& $91.6\pm15.5$ & $84.3\pm19.7$ & $75.8\pm17.6$ & $56.8\pm11.8$ & $55.8\pm10.4$ & $54.9\pm8.8$  \\
        \bottomrule
    \end{tabular}
    \vspace{-0.21cm}
\end{table*}

\section{Discussion}\label{sec:discussion}
\subsection{Training}

\fref{fig:curves} shows how the mean episode rewards evolve over training steps across five training runs for 
the two methods (RL-DWA and pure RL) with the different sensing configurations ($N=15,25,50$).
The shaded areas around each curve represent the standard deviations, indicating variability in performance. The proposed \rldwa outperforms the corresponding RL baseline in all the sensing conditions, confirming the benefit of combining reinforcement learning with a DWA-based planner.

In particular, the \rldwa with both $N=50$ and $N=25$ shows a rapid increase in reward during the initial training phase, converging towards high values of $5657 \pm 344$ and $5278 \pm 438$, respectively. 
Despite the comparable final rewards achieved, their variance profiles reveal an important difference. Specifically, the configuration with $N=50$ exhibits narrower intervals in the training stages, suggesting more consistent convergence across runs. This indicates that a higher sensing resolution not only improves average performance but also enhances reproducibility, reducing the sensitivity of the learning process to random initialization and exploration variability.
The \rldwa ($N=15$) variant stabilizes at a significantly lower level ($3285 \pm 462$), indicating that an overly sparse perception penalizes the planner by limiting the quality of the navigation.
Quantitatively, this trend is confirmed by the AUC, which reaches $\num{2.12e6} \pm \num{0.23e6}$ for $N=50$ and $\num{1.98e6} \pm \num{0.23e6}$ for $N=25$, compared to only $\num{1.17e6} \pm \num{0.21e6}$ for $N=15$. The higher AUC values demonstrate both faster and more stable accumulation of reward when richer sensing is available.

In 
contrast, the RL baselines 
achieve markedly lower performance. After $\num{1e6}$ training steps, the 
$N=50$ configuration reaches only $1806 \pm 1161$, while the lower-resolution variants ($N=25$ and $N=15$) attain $1471 \pm 932$ and $577 \pm 225$, respectively. 
Moreover, reward growth is slower and variability across runs remains high, especially for $N=50$ and $N=25$, indicating unstable learning dynamics.
The apparent stability of the $N=15$ configuration instead arises from uniformly poor performance, as all runs plateau at low reward values.

Overall, 
integrating the DWA planner substantially improves both convergence speed and final performance compared to pure RL. 
Sensing resolution plays also a critical role: while very sparse perception ($N=15$) significantly limits navigation capabilities, intermediate resolution ($N=25$) already captures most of the benefits, and higher resolution ($N=50$) provides incremental gains together with improved reproducibility.

\subsection{Test}

\fref{fig:training_res} presents the performance of the different controllers in known navigation scenarios, highlighting the impact of sensor resolution and control strategies on path completion $p_{\%}$ and occupied volume $V_{\%}$. 
Among the proposed methods, the \rldwa controllers achieve the best trade-off between navigation reliability and adaptive deformation. 
The $N=50$ configuration achieves the most robust navigation, with path completion rates consistently close to 100$\%$ (Mdn: 99.28$\%$) and narrow variability, demonstrating high reproducibility. The $N=25$ configuration delivers a comparable level of deformation to $N=50$ (both around 57$\%$), showing that moderate sensing is sufficient for effective shape adaptation, although path completion is less reliable (Mdn: 79.26$\%$) and more variable. The $N=15$ case still provides acceptable deformation (Mdn: 49.87$\%$), but completion drops (Mdn: 73.28$\%$) and dispersion increases, confirming that overly sparse sensing reduces navigation robustness.
By contrast, the pure RL baselines perform consistently worse. Even with dense sensing ($N=50$), path completion remains limited (Mdn: 57.06$\%$) and highly variable, while $N=25$ and $N=15$ degrade further, reaching 44.83$\%$ and 26.81$\%$, respectively. Regarding deformations, RL ($N=50$) achieves values comparable to RL-DWA (Mdn: $61.32\%$).
At lower resolutions, deformation capability is severely reduced (Mdn: 31.56$\%$ at $N=25$ and 17.93$\%$ at $N=15$). This degradation arises because sparse sensing compromises navigation abilities, causing the robot to approach or collide with the vessel boundaries more frequently. As a result, volume occupation is not optimized, since deformation is limited to avoid collisions.
The model-based controller ($N=50$), instead, achieves relatively high deformation (Mdn: 60.36$\%$) but 
limited navigation success (Mdn: 43.77$\%$), 
reflecting the limitations of the underlying graph-based method, which is sufficiently accurate to infer cross-sectional geometry but less effective at estimating the central path of the vessel. 
%
%
At lower sensing resolutions ($N=25$ and $N=15$), sparse point clouds prevent reliable Delaunay/Voronoi graph construction, leading to poor vessel centerline estimation and degraded deformation control.

In unseen scenarios (\fref{fig:test_res}), 
the effect of sensor resolution becomes more evident. The \rldwa ($N=50$) generalizes best, with near-perfect path completion (Mdn: 99.22$\%$) and stable deformation (Mdn: 53.22$\%$), showing strong robustness to novel environments. The $N=25$ variant maintains comparable deformation (Mdn: 52.8$\%$) but with reduced completion (Mdn: 75.22$\%$) and larger variability, indicating occasional failures. The $N=15$ configuration generalizes poorly: deformation remains non-negligible (Mdn: 50.98$\%$), but completion decreases further (Mdn: 71.37$\%$) and variance is high. Pure RL degrades drastically, with completion dropping to 45.75$\%$ at $N=50$ and even lower at $N=25$ and $N=15$, accompanied by very limited deformation capability (25.33$\%$ and 14.01$\%$, respectively). The model-based controller with $N=50$ performs similarly to the previously discussed scenarios, retaining high deformation (Mdn: 51.06$\%$) 
but showing limited path completion (Mdn: 44.14$\%$), and consistently achieving zero-score at lower resolutions. 

\subsection{Inference time}
The results reported in \tref{tab:inference_times} confirm that our approach 
achieves fast inference, with mean per-step times consistently below 2~ms, ensuring suitability for real-time deployment. 
The RL module inference time is constant and it does not depend on the input size $N$. 
Conversely, the DWA module inference time increases with $N$, as the cost $G$~\eref{eq:dwa_costfunction} is evaluated for each admissible velocity in~\eref{eq:velocity_window}, and its clearance term scans all points. 
If the planner explores a $k$-dimensional velocity space (with $k=3$ in our case) discretized into $M$ samples per dimension, the total number of velocity candidates is $M^{k}$. For each candidate, the trajectory is sampled at $P = \tfrac{T}{\Delta t}$ points over the fixed horizon $T$. The resulting per-cycle time complexity is therefore $O(M^{k} N P)$.
In~\eref{eq:dwa_costfunction}, the terms $vel(\cdot)$, $dir(\cdot)$, and $head(\cdot)$ are $O(1)$ per candidate, while $clear(\cdot)$ dominates due to its $O(N)$ dependence. For small $N$, fixed overheads and the per-candidate $O(1)$ terms mask the linear trend, so the measured time may appear nearly unchanged across different $N$.


\subsection{Robustness under uncertainties}

%
%

The robustness analysis reported in \tref{tab:exp_robustness} highlights how increasing sensor noise impacts path completion and occupied volume differently across sensing resolutions.

For path completion, all controllers degrade with noise, but the slope of degradation varies. With $N=15$, performance drops steeply from $72.1 \pm 22.0\%$ 
to $51.2 \pm 27.1\%$, 
a loss of nearly 30$\%$. The $N=25$ configuration shows the mildest decline (
about $15\%$), indicating comparatively stronger robustness. The $N=50$ case experiences a larger  drop than $N=25$ (about $17\%$), 
yet it still preserves the highest overall completion rates across all noise levels. 

For occupied volume, the effect of noise is much less pronounced. The variations remain below $4\%$ for all resolutions. 
This stability indicates that deformation control is inherently more robust to sensor perturbations than navigation, with higher resolutions ($N=50$) showing slightly more resilience.

\section{Conclusion and Future Work}\label{sec:conclusion}
This work presented a novel hybrid navigation framework that combines RL with DWA for adaptive 3D local navigation of high-DoF 
robots under sparse sensory data conditions. 
Our method 
integrates learning-based motion adaptation with real-time velocity optimization, proving effective in achieving adaptive navigation in constrained 3D environments, with low inference times. 
Experimental results show that our approach consistently outperforms both pure RL and model-based baselines in path completion and deformation control. By combining the adaptability of RL with the structured guidance of DWA, the proposed method achieves strong generalization and robustness to sensor noise, especially at higher sensing resolutions, while preserving reliable deformation control under sparse sensing. In contrast, pure RL suffers from unstable policies and poor generalization, whereas the model-based controller lacks robustness under sparse sensory conditions.


From a safety perspective, the DWA component acts as a structured filter between the learned policy and the executed translational command. Rather than directly commanding the full robot velocity, the RL policy adapts the DWA cost-function weights, while the planner selects the final linear velocity from admissible candidates. 
	
Although the framework was evaluated in a simplified simulation scenario, its structure is general and can be transferred to real robotic platforms by redefining the system-specific dynamic window, actuation bounds, sensing model, and safety constraints. The same high-level architecture could also be adapted to macroscopic systems operating in confined 3D spaces, such as micro-UAVs or coordinated drone swarms navigating pipes, tunnels, or similar constrained environments.

Future work will focus on transferring the approach to real-world applications. Specifically, we plan to adapt the control strategy to a magnetically actuated robot, leveraging recent advances that allow coupled control of motion and deformation in ferrofluidic microrobots~\cite{Xinjian2023Combined}. We will replace the simulated laser-scan input with 3D reconstructions from imaging modalities~\cite{Pane2021Realtime,li2020oct} 
which will also provide estimates of the robot’s pose and deformation. Furthermore, we aim to integrate this local framework with global path planning methods to automatically generate waypoint sequences in complex environments, enabling end-to-end navigation that combines global optimality with local adaptability. 

\bibliographystyle{IEEEtran}
\bibliography{biblio}
\end{document}